\documentclass{article}
\usepackage{arxiv}

\usepackage[utf8]{inputenc} 
\usepackage[T1]{fontenc}    
\usepackage{hyperref}       
\usepackage{url}            
\usepackage{booktabs}       
\usepackage{amsfonts}       
\usepackage{nicefrac}       
\usepackage{microtype}      
\usepackage{lipsum}		
\usepackage{graphicx}
\usepackage{natbib}
\usepackage{doi}
\usepackage{graphicx}
\usepackage{multirow}
\usepackage{amsmath}
\usepackage{listings}
\usepackage{diagbox}
\def\ELp{$\mathcal{EL}^+$}

\def\ER{$\mathcal{ER}$}

\title{On the Capabilities of Pointer Networks for Deep Deductive Reasoning}

\author{ \href{https://orcid.org/0000-0000-0000-0000}{\includegraphics[scale=0.06]{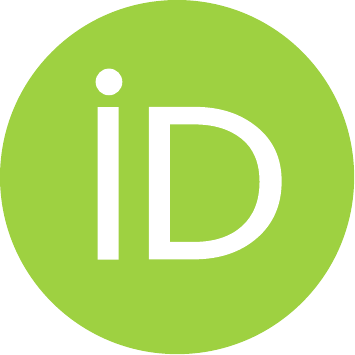}\hspace{1mm}Monireh Ebrahimi}
		\\
	Department of Computer Science\\
	Kansas State University\\
	\texttt{monireh@ksu.edu} \\
	\And
	\href{https://orcid.org/0000-0000-0000-0000}{\includegraphics[scale=0.06]{orcid.pdf}\hspace{1mm}Aaron Eberhart} \\
	Department of Computer Science\\
	Kansas State University\\
	\texttt{aaroneberhart@ksu.edu} \\
		\And
	\href{https://orcid.org/0000-0000-0000-0000}{\includegraphics[scale=0.06]{orcid.pdf}\hspace{1mm}Pascal Hitzler} \\
	Department of Computer Science\\
	Kansas State University\\
	\texttt{hitzler@ksu.edu} \\
}

\renewcommand{\shorttitle}{\textit{Monireh Ebrahimi et al.} }
\hypersetup{
pdftitle={A template for the arxiv style},
pdfsubject={q-bio.NC, q-bio.QM},
pdfauthor={David S.~Hippocampus, Elias D.~Striatum},
pdfkeywords={First keyword, Second keyword, More},
}

\begin{document}
\maketitle

\begin{abstract}
The importance of building neural networks that can learn to reason has been well recognized in the
neuro-symbolic community. In this paper, we apply neural pointer networks for conducting reasoning over symbolic knowledge bases. In doing so, we explore the benefits and limitations of encoder-decoder architectures in general and pointer networks in particular for developing accurate, generalizable and robust neuro-symbolic reasoners. Based on our experimental results, pointer networks performs remarkably well across multiple reasoning tasks while outperforming the previously reported state of the art by a significant margin. We observe that the Pointer Networks preserve their performance even when challenged with knowledge graphs of the domain/vocabulary it has never encountered before. 
To the best of our knowledge, this is the first study on neuro-symbolic reasoning using Pointer Networks. We hope our impressive 
results on these reasoning problems will encourage broader exploration of pointer networks' capabilities for reasoning over more complex logics and for other neuro-symbolic problems.
\end{abstract}

\section{Introduction}
\label{intro}
The study of architectures and methods for artificial neural networks so that they can learn and perform tasks from the realm of logic-based knowledge representation and reasoning has a long-standing tradition \cite{besold2017neural}. This research area is sometimes referred to as ``neuro-symbolic integration'' (or ``neural-symbolic integration'') and there are at least two primary rationales that can be found in the literature on the subject. The first is the desire to arrive at systems that combine the robustness and trainability of artificial neural networks with the transparency and interpretability of knowledge-based systems, while at the same time making use of structured background knowledge. The second rationale is more prevalent in cognitive science and lies in addressing the fundamental gap between symbolic and subsymbolic representation and processing, based on the observation that humans perceive much of their own thinking, introspectively, as symbolic, while the physical structure of the brain gives rise to artificial neural networks as a mathematical and computational abstraction.

Many of the earlier lines of research on neuro-symbolic integration, discussed primarily from a cognitive science perspective, can be found in \cite{besold2017neural}. Of particular interest is the integration of deep learning with logics that are not propositional in nature, since propositional logic is of limited applicability to knowledge representation and reasoning tasks. In the wake of deep learning breakthroughs, fundamental issues around neuro-symbolic integration have recently received increased attention with some progress being made as new approaches emerge. In particular, there has been progress in developing neural networks that can learn to reason. These include the Neural Theorem Prover (NTP) and its variations \cite{rocktaschel2017end,DBLP:conf/akbc/RocktaschelR16,DBLP:conf/icml/Minervini0SGR20,DBLP:conf/aaai/MinerviniBR0G20}, Logic Tensor Networks (LTN) \cite{serafini2016learning,DBLP:conf/aaaiss/BianchiH19,DBLP:journals/corr/abs-2012-13635}, and the application of memory networks and LSTMs \cite{ebrahimi-apin-2021} and others \cite{MakniH19,HoheneckerL20}. Yet, there is still much work to do in terms of new model development and investigation of inductive bias in existing architectures and their reasoning capability.

This paper tries to fill this gap by examining the reasoning capability of pointer networks \cite{DBLP:conf/nips/VinyalsFJ15} in emulating deductive reasoning. 
Pointer Networks and their variations have been applied successfully to a variety of sophisticated tasks including theoretical computer science problems (i.e., NP-hard Travelling Salesman Problem (TSP), Delaunay Triangulation, and Convex Hull) \cite{DBLP:conf/nips/VinyalsFJ15} and practical problems like abstractive \cite{DBLP:conf/acl/SeeLM17} and extractive \cite{DBLP:conf/acl/JadhavR18} text summarization, code completion \cite{DBLP:conf/ijcai/LiWLK18}, and dependency parsing \cite{DBLP:conf/acl/Fernandez-Gonzalez20a,DBLP:conf/naacl/Fernandez-Gonzalez19,DBLP:conf/acl/HovyMNPHL18}. Nevertheless, almost nothing is known about their potential for conducting logical deductive reasoning accurately. In fact, they have been mainly used for solving discrete combinatorial optimization problems because of their variable-size output vocabulary and for resolving the rare or out-of-vocabulary problem in Natural Language Processing. We hypothesize that using the pointer attention to decide what elements of the input knowledge base should be chosen as the output, and in which order, will work well for deductive reasoning tasks. Indeed, using pointer networks we can mimic human reasoning behaviour where one can learn to choose a set of symbols in different locations and copy these symbols to suitable locations to generate new logical consequences based on a set of predefined logical entailment rules. To verify this, here, we explore the capabilities and limitations of pointer networks for performing deductive reasoning on Resource Description Framework (RDF) \cite{rdfs} and \ELp{} \cite{BaaderBL05} knowledge bases in terms of accuracy, and generalizability.
Based on our experimental results, pointer networks perform remarkably well across multiple reasoning tasks while outperforming the previously reported state of the art by a significant margin. We observe that the Pointer Networks preserve  their  performance  even  when  challenged  with  knowledge  graphs  of the  domain/vocabulary  it  has  never  encountered  before. To our knowledge, this work is the first attempt to reveal the impressive power of pointer networks for conducting deductive reasoning.

In terms of the logic, in this paper we are looking at two logics with different expressivity, power, and reasoning difficulty. The Resource Description Framework Schema (RDFS) \cite{rdfs} is non-trivial (and non-propositional), yet one of the simplest widely used logics: It is a mature W3C Semantic Web standard that is commonly used to express knowledge graphs and linked data \cite{cacm-swsurvey}, and many corresponding data sets are freely available on the World Wide Web~\cite{RietveldBHS17}. The standard carries a model-theoretic semantics which defines deductive entailment \cite[Section 9.2]{rdfsemantics}, and reasoning over RDFS is usually done using rule-based reasoning engines. The second logic is the description logic \ELp{} (or \ER) \cite{BaaderBL05} that is the basis for the W3C standard OWL EL \cite{owl2-primer}. It is considered to be a rather inexpressive but practically useful logic \cite{SchulzSBB09} and generally used for expressing ontologies and knowledge graph schemas \cite{FOST} particularly in medical domain ontologies. 

In short this paper strives to answer two main research questions: “Can Pointer Networks perform logical deductive reasoning using pointer attention?”, and more generally, “Can other attention-based sequence-to-sequence models like self-attention based popular Transformer architectures successfully perform the same task?”,``How well do pointer network reasoners perform  on completely new knowledge graphs?", and finally, ``How robust is our model to noise?". To answer these questions we conduct a set of experiments by applying pointer networks and transformers to RDFS and \ELp{} reasoning tasks. We believe, the answer to our third question is particularly very important since it a very big step toward developing accurate yet symbol-invariant deep deductive reasoners which generalize very well on unseen knowledge bases of differing domain or vocabulary.
The contributions of this work are fourfold:
\begin{enumerate}
  \item A novel paradigm for viewing a symbolic reasoning problem as a pointing problem.
  \item Pointer Networks are used to neurally resolve symbolic reasoning for the first time.
  \item The proposed approach is able to transfer its reasoning ability to new domain/vocabulary knowledge graph of same logic.
  \item We report the state-of-the-art performance of the \ELp{} and RDF reasoning.
\end{enumerate}

The remainder of the paper is organized as follows. In Section~\ref{sec:2} we discuss related research efforts, 
more precisely an overview of recent work on deep deductive reasoning over RDFS, \ELp{}, and other logics,
followed by a list of various tasks where pointer networks have been effectively applied. In Section~\ref{sec:3}, we concretely present the deep learning architecture and the logics we used. In Sections~\ref{sec:4}, we present an experimental evaluation of our approach and discuss our findings. We conclude and discuss future work in Section~\ref{sec:5}.

\section{Related Work}
\label{sec:2}
\subsection{Deep Deductive Reasoning}
\label{sec:2-1}
Training artificial neural networks to learn deductive reasoning is a hard machine learning task that was out of reach before the advent of deep learning. In the last few years, several publications have shown that deep deductive reasoning -- using deep learning methods -- is possible. We will briefly review the core body of existing work. As we will see, it remains a hard task, even for deep learning. 

Before we do so, though, let us point out that our work is different from what is usually called knowledge graph completion, or the study of knowledge graph embeddings, although we deal with logics relevant for knowledge graphs~\cite{cacm-swsurvey}: Knowledge graph completion (sometimes called link prediction or knowledge graph refinement) \cite{DBLP:journals/semweb/Paulheim17} refers to enriching a knowledge graph with additional relationships that are statistically induced, sometimes using machine learning methods. In contrast to this, we are studying \emph{deductive} reasoning, which is not based on statistics or likelihood, but based on a mathematical, logical calculus that derives additional statements which were already implicit -- in a mathematically precisely defined sense -- in the statements already made. Deductive inference tasks are usually hard computationally (e.g., for propositional logic, it is NP-complete), and are traditionally addressed using complex but provably correct algorithms -- correctness in this sense is in relation to the underlying mathematical definitions that determine what is, and what is not, a deductive logical consequence. The study of knowledge graph embeddings~\cite{DBLP:journals/semweb/RistoskiRNLP19}, in isolation, is about the learning of representations of knowledge graphs in multi-dimensional Euclidean space. While embeddings are often a component of deep deductive reasoning systems, our goal is the overall functionality of deep deductive reasoning, and not just knowledge graph embeddings in isolation.

A good overview of existing deep deductive reasoning work is \cite{ebrahimi-apin-2021}. It appears to be appropriate to distinguish between the different logics that are addressed in the literature, the reasonable assumption being that less complex logics are easier to learn, and this resonates with the as yet limited body of work. We refer to \cite{FOST} for background on all the mentioned logics. We know about investigations of RDFs \cite{DBLP:conf/aaaiss/EbrahimiSBXEDKH21,MakniH19}, of \ELp \cite{DBLP:conf/aaaiss/EberhartEZSH20}, of OWL RL \cite{HoheneckerL20}, and of first-order predicate logic (FOL) \cite{DBLP:conf/aaaiss/BianchiH19}. 

\begin{table}[ht]
\begin{tabular}{r|c|c|c|c|r}
     paper & logic & transfer & generative & scale & performance \\
     \hline
 \cite{DBLP:conf/aaaiss/EbrahimiSBXEDKH21}    & RDFS & yes & no & moderate & high\\
 \cite{MakniH19} & RDFS & no & yes & low & high\\
 \cite{DBLP:conf/aaaiss/EberhartEZSH20} & \ELp & yes & yes & moderate & low\\
 \cite{HoheneckerL20} & OWL RL & no* & no & low & high \\
 \cite{DBLP:conf/aaaiss/BianchiH19} & FOL & no & yes & very low & high
\end{tabular}\\\\
\caption{Overview of published deep deductive reasoning work. See the main text for details on the columns. no* indicates that the paper claims that transfer is possible in principle, but it was not demonstrated or evaluated to what extent transfer really happens.}\label{tab:soa}
\end{table}

We give an overview of the key aspects of each of these in Table \ref{tab:soa} -- we admit that some interpretations in this table may be somewhat subjective. The column ``transfer" indicates whether the system was demonstrated to have a good transfer capability to previously unknown and very different knowledge bases. The column ``generative" indicates whether the system generates all (under certain finiteness constraints) deductive inferences in one run -- if not, then it would usually be query-based, i.e. it would be able to tell whether a given logical expression is a logical consequence of the knowledge base. The column ``scale" indicates how large the input knowledge bases were in the experiments, ranging from a few logical statements as in the FOL case to RDF graphs with 1,000 triples in \cite{DBLP:conf/aaaiss/EbrahimiSBXEDKH21}. The column ``performance" indicates how well the system learned to reason; ``high" indicates 70\%{} or more in terms of f-measure, while ``low" indicates values just a bit better than random guessing. 

For a deep deductive reasoner, we would ideally like to have it on an expressive logic, with transfer, generative, at massive scale, and with high performance. For all the referenced works, except the FOL one, the scale aspect has not been systematically explored yet; for the FOL case, it does look rather unfavorable as discussed in \cite{DBLP:conf/aaaiss/BianchiH19}. Otherwise, it is important to note that only one of the works is both generative and able to transfer, however this was also the system with very low performance. As we will see, our new approach we report on in this paper is able to do transfer and is generative, with high performance. This is the key contribution of this paper.


\subsection{Pointer Networks}\label{sec:2-4}
Pointer Networks and their variations have been applied successfully to a variety of sophisticated tasks including theoretical computer science problems (i.e., NP-hard Travelling Salesman Problem (TSP), Delaunay Triangulation, and Convex Hull \cite{DBLP:conf/nips/VinyalsFJ15} as well as 0–1 Knapsack problem \cite{DBLP:conf/icaci/GuH18}) and practical problems like abstractive \cite{DBLP:conf/acl/SeeLM17} and extractive \cite{DBLP:conf/acl/JadhavR18} text summarization, code completion \cite{DBLP:conf/ijcai/LiWLK18}, dependency parsing \cite{DBLP:conf/acl/Fernandez-Gonzalez20a,DBLP:conf/naacl/Fernandez-Gonzalez19,DBLP:conf/acl/HovyMNPHL18}, named entity boundary detection \cite{DBLP:conf/ijcai/0034YS19}, conversation disentanglement\cite{DBLP:conf/emnlp/YuJ20}, anaphora resolution \cite{DBLP:journals/prl/LeeJP17}, paragraph ordering \cite{DBLP:journals/nn/PandeyC20}, paraphrase generation for data augmentation \cite{DBLP:conf/iccS/GuptaK20}, entity linking \cite{DBLP:conf/semweb/BanerjeeCDL20}, and airline itinerary prediction \cite{DBLP:conf/kdd/MottiniA17}. Nevertheless, almost nothing is known about their possible application and ability for conducting logical deductive reasoning accurately. In fact, they have been mainly used for solving discrete combinatorial optimization problems because of their variable-size output vocabulary and for resolving the rare or out-of-vocabulary problem in Natural Language Processing. 

\section{Methodology}
\label{sec:3}
In order to explain more formally what we are setting out to do, let us first re-frame our entailment problem as an input-output mapping task: Given some logic $\mathcal{L}$, for each theory $T$ over $\mathcal{L}$, the set $c(T)=\{F\mid T\models_\mathcal{L} F\}$ of all formulas over $\mathcal{L}$ that are entailed by $T$; we call $c(T)$ the \emph{completion} of $T$. We can then attempt to train a neural network to produce $c(T)$ for any given $T$ over $\mathcal{L}$, i.e., we would use pairs $(T,c(T))$ as input-output training pairs for a generative deep deductive reasoner.

\subsection{Logics}
\label{sec:3-1}
\paragraph{RDF}
The Resource Description Framework RDF, which includes RDF Schema (RDFS) \cite{rdf-spec,FOST} is an established and widely used W3C standard for expressing knowledge graphs. The standard comes with a formal semantics\footnote{In fact, it comes with three different ones, but we have only considered the most comprehensive one, the RDFS Semantics.} that defines an entailment relation. An RDFS knowledge base (KB) 
is a collection of statements stored as \emph{triples} $(e1,r,e2)$ where $e1$
 and $e2$ are called \emph{subject} and \emph{object}, respectively, while $r$ is a binary relation between $e1$ and $e2$. In the context of RDF/RDFS, the triple notation $(e1,r,e2)$ is more common than a notation like $r(e1,e2)$ as it is suggestive of a node-edge-node piece of a labelled graph, and so we will use the triple notation.
 
 As a logic, RDFS is of very low expressivity and reasoning algorithms are very straightforward. In fact, there is a small set of thirteen entailment rules~\cite{world2014rdf}, fixed across all knowledge graphs, which are expressible using Datalog.\footnote{Datalog is equivalent to function-free definite logic programming \cite{seda-book}.} These thirteen rules can be used to entail new facts. 
 

 
\begin{table}[ht]
\caption{Selected RDFS Completion Rules}\label{tab:rdfcomp}	
\begin{align}
		(x, \text{rdfs:subClassOf}, y), (y,\text{rdfs:subClassOf}, z) &\models (x,\text{rdfs:subClassOf}, z) \label{rdfs:sCo-trans} \\
				(x, \text{rdfs:subPropertyOf}, y), (y,\text{rdfs:subPropertyOf}, z) &\models (x,\text{rdfs:subPropertyOf}, z) \label{rdfs:sPo-trans} \\
				(x, \text{rdfs:subClassOf}, y),(z,\text{rdf:type}, x) &\models (z, \text{rdf:type}, y) \label{rdfs:sCo-type} \\
				(a, \text{rdfs:domain}, x),(y, a, z) &\models (y, \text{rdf:type}, x) \label{rdfs:domtype} \\
				(a, \text{rdfs:range}, x),(y, a, z) &\models (z, \text{rdf:type}, x) \label{rdfs:rangetype} 
	\end{align}
\end{table}

 Table \ref{tab:rdfcomp} shows examples for some of these entailment rules. The identifiers $x,y,z,a$ are variables. The remaining elements of the triples are pre-fixed with the rdfs or rdf namespace (a concept borrowed from XML) and carry a specific meaning in the formal semantics of RDFS. E.g., rdfs:subClassOf indicates a sub-class (or sub-set) relationship, i.e. Rule \ref{rdfs:sCo-trans} states transitivity of the rdf:subClassOf binary relation. Likewise, in Rule~\ref{rdfs:sPo-trans}, 
 $(x, \text{rdfs:subPropertyOf}, y)$ indicates that $x,y$ are to be understood as binary relations, where $x$ is a restriction (called a \emph{subproperty}) of $y$. In Rule \ref{rdfs:sCo-type}, the triple $(z, \text{rdf:type}, x)$ indicates that $z$ is a member of the class (or set) $x$. In Rules \ref{rdfs:domtype} and \ref{rdfs:rangetype}, rdfs:domain and rdfs:range indicate domain respectively range of $a$, which is to be interpreted as a binary relation. The rules are applied exhaustively on an input RDF knowledge base, i.e. inferred triples are added and then rule execution continues taking the new triples also into account.
 
\paragraph{\ELp}
The standard reasoning task over \ELp{} is called \emph{classification}  and can be understood as the computation of all formulas of the form $\forall x (p(x) \to q(x))$ entailed by the given theory, and the set of all these formulas, which is called the \emph{completion} of the input theory, is finite if the input theory is finite. 

Formally, let $N_C$ be a set of atomic classes (or concepts, or class names), let
$N_R$ be a set of roles (or properties), and let $N_I$ be a set of individuals. Complex class expressions (or simply complex classes or classes) in the description logic \ELp{}
are defined by the grammar
$$C ::= A|C_1 \sqcap C_2|\exists R.C,$$
where $A \in N_C$, $R \in N_R$, and $C_1$, $C_2$, and $C$ are complex class expressions. A TBox in \ELp{}
is a set of general class inclusion axioms (or TBox statements) of the form $C \sqsubseteq D$, where
$C$, $D$ are (complex) classes. We use $C \equiv D$ as abbreviation for the two statements $C \sqsubseteq D$
and $D \sqsubseteq C$. An RBox in \ELp{} is a set of general role inclusion axioms (or RBox statements)
of the form $R_1 \circ \cdot \cdot \cdot \circ R_n \sqsubseteq R$, where $R$, $R_i \in N_R$ (for all $i$). 
An \ELp{} knowledge base (or ontology) is a set of TBox and RBox statements. 

\ELp{} is in fact a fragment of first-order predicate logic: all statements can be translated into it, and the inherited semantics is exactly the first-order predicte logic semantics -- details can be found in \cite{FOST}. Classification is known to be P-complete. 

An \ELp{} \emph{normal form} knowledge base contains only axioms of the following forms.
\begin{align*}
    C &\sqsubseteq D & C_1\sqcap C_2 &\sqsubseteq D & C &\sqsubseteq \exists R.D \\
    \exists R.C &\sqsubseteq D & R_1 &\sqsubseteq R & R_1\circ R_2 &\sqsubseteq R
\end{align*}
As usual, every \ELp{} knowledge base can be cast into normal form in polynomial time, and such that it suffices to perform classification over the normal form knowledge base. 

Given an \ELp{} knowledge base $K$ in normal form, the completion $\text{comp}(K)$ of $K$ can for example be obtained
from $K$ by exhaustively applying the completion rules from Table \ref{tab:ercomp}. There are of course different ways to perform classification using reasoning algorithms, e.g. reasoning can also be encoded using a larger number of Datalog rules 
which remain fixed across input theories \cite{DBLP:conf/ijcai/Krotzsch11}. So on the surface
this seems similar to RDFS reasoning. However \ELp{} as a logic has a different look and feel: RDFS
reasoning focuses on the derivation of new facts from old facts, while \ELp{} is about the processing of
schema knowledge, in particular subclass relationships, in the presence of existential quantification (which is completely absent from RDFS). The transformation into Datalog reasoning is also more complicated than for RDFS.

\begin{table}[ht]
\caption{\ELp{} Completion Rules}
\label{tab:ercomp}	\centering
	\begin{tabular}{lccccccl}\hline
		(1)&$A \sqsubseteq C $& &$C \sqsubseteq D$& &&$\models$&$ A \sqsubseteq D$\\
		(2)&$A \sqsubseteq C_1$& &$A \sqsubseteq C_2$& &$C_1 \sqcap C_2 \sqsubseteq D$&$\models$&$A \sqsubseteq D$\\
		(3)&$A \sqsubseteq C$& &$C \sqsubseteq \exists R.D$& &&$\models$&$A \sqsubseteq \exists R.D$\\
		(4)&$A \sqsubseteq \exists R.B$& &$B \sqsubseteq C$& &$\exists R.C \sqsubseteq D$&$\models$&$A \sqsubseteq D$\\
		(5)&$A \sqsubseteq \exists S.D$& &$S \sqsubseteq R$& &&$\models$&$A \sqsubseteq \exists R.D$\\
		(6)&$A \sqsubseteq \exists R_1.C$& &$C \sqsubseteq \exists R_2.D$& &$R_1 \circ R_2 \sqsubseteq R$&$\models$&$A \sqsubseteq \exists R.D$\\\hline
	\end{tabular}	
\end{table}

%

\subsection{Pointer Networks}
\label{sec:3.2}
\begin{figure}[t]
   \centering
   \includegraphics[width=\textwidth]{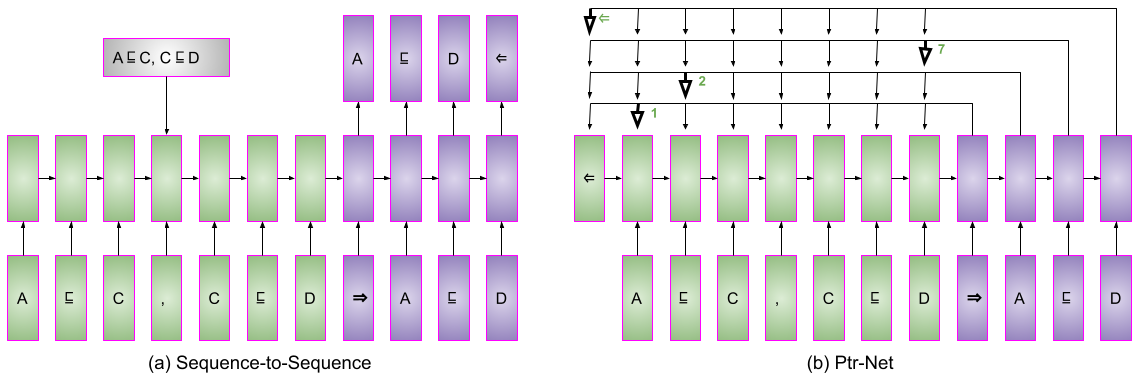}
   \caption{\ELp\ Completion Rule (1): "(a) Sequence-to-Sequence - An RNN (green) processes the input sequence to create a code
vector that is used to generate the output sequence (purple) using the probability chain rule and
another RNN. 
(b) Ptr-Net - An encoding RNN converts the input sequence
to a code (green) that is fed to the generating network (purple). At each step, the generating network
produces a vector that modulates a content-based attention mechanism over inputs. The
output of the attention mechanism is a softmax distribution with dictionary size equal to the length
of the input."}
   \label{fig:pointer}
\end{figure}
Pointer networks is an encoder-decoder architecture based model which uses attention as a pointer to choose an element of the input in each decoding time step. The remarkable main advantage of this model compared to other sequence-to-sequence models like Transfromers is that the learned models generalize beyond the maximum lengths that they were trained on. Thus, they were initially proposed to generate a correct variable size output sequence, given an input sequence consisting of a variable size combinatorial optimization problem. Amazingly, it has even outperformed the fixed input size problem baseline, showing its potential to be used in wider applications.

Inspired by the above-mentioned advantages, here we have used pointer networks to copy the symbols from the knowledge base via pointing to generate logical consequences.

In our corpus, each knowledge graph and its completion denoted as $(T,c(T))$ comprises the sequence of symbols. Given the $(T,c(T))$ pair, we are feeding the pointer network with $(\mathcal{T},\mathcal{C}(\mathcal{T})^\prime)$ where $\mathcal{T} = \{T_1,...,T_n\}$ is a sequence of n symbols each refer to an element in our input knowledge graph and $\mathcal{C}(\mathcal{T})^\prime = \{c(T_1)^\prime,...,c(T_m)^\prime\}$ is a sequence of $m$ indices each between $1$ and $n$.
The sequence-to-sequence model then computes the conditional probability $p(c(T_i)^\prime|c(T_1)^\prime,...,c(T_{i-1})^\prime, \mathcal{T})$ changing the Bahdanau \cite{DBLP:journals/corr/BahdanauCB14} attention to the pointer attention as follows:
\begin{gather}
\label{eq:1}
u^i_j = v^T \tanh(W_1e_j + W_2d_i) \qquad\text{for } j \in(1,....,n) \qquad\text{ and}\\
\label{eq:2}
p(c(T)^\prime_i|c(T)^\prime_1,...,c(T)^\prime_i-1, \mathcal{T}) = \text{Softmax}(u^i),
\end{gather}
where $(e_1,...,e_n)$ and $(d_1,...,d_m)$ denote the encoder and decoder hidden states respectively, and $v$, $W_1$, and $W_2$ are learnable parameters of the output model. The softmax 
 $$\text{Softmax}(u_{i}) = \frac{e^{(u_{i})}}{\sum_{j}e^{(u_{j})}}$$
 normalizes the vector $u^i$ of length $n$ to be an output distribution over the dictionary of inputs. Indeed the model uses $u^i_j$ as pointers to the input symbols.

\section{Experimental Setup}
\label{sec:4}
In this section, we describe the detail of datasets, training process, the baseline models, and the performance of our proposed model in terms of correctness, and generalizability
in comparison to existing baselines.
\subsection{Datasets}
\label{sec:4-1}
We evaluate different approaches on two benchmarked
datasets:
\paragraph{\ELp Dataset}
To provide sufficient training input to our network we followed the same synthetic generation procedure as proposed in \cite{DBLP:conf/aaaiss/EberhartEZSH20} that combines a structured forced-lower-bound reasoning sequence with a connected
randomized knowledge base. This allows us to rapidly generate many normal semi-random \ELp knowledge bases of
arbitrary reasoning difficulty. For this experiment we choose the knowledge bases of size 40, 50, and 120 statements with
a moderate difficulty setting so that it can compare with nonsynthetic data. To ensure that the randomized statements do not shortcut this pattern, the random
statements are generated in a nearly disjoint space and connected only to the initial seed term. This ensures that at least
one element of the random space will also produce random
entailments for the duration of the sequence, possibly longer.
Our procedure also guarantees that each completion rule will
be used at least once every iteration of the sequence so that
all reasoning patterns can potentially be learned by the system.
An example of a graph and its corresponding inference in our \ELp dataset is demonstrated in Table \ref{tab:elp-sample}.
\begin{table}[t]
\centering
{
\begin{tabular}{|ll|}
\hline
& $ C0 \sqsubseteq C0 $ \\ 
& $ C1 \sqsubseteq C1 $ \\ 
& $ C2 \sqsubseteq C2 $ \\ 
& $ C3 \sqsubseteq C3 $ \\ 
& $ C4 \sqsubseteq C4 $ \\ 
& $ C5 \sqsubseteq C5 $ \\ 
& $ C6 \sqsubseteq C6 $ \\ 
& $ C7 \sqsubseteq C7 $ \\ 
& $ C8 \sqsubseteq C8 $ \\ 
& $ C9 \sqsubseteq C9 $ \\ 
& $ C10 \sqsubseteq C10 $ \\ 
& $ C11 \sqsubseteq C11 $ \\ 
& $ C12 \sqsubseteq C12 $ \\ 
& $ C13 \sqsubseteq C13 $ \\ 
& $ C2 \sqsubseteq C9 $ \\ 
& $ C4 \sqsubseteq C2 $ \\ 
& $ C8 \sqsubseteq C0 $ \\ 
& $ C9 \sqsubseteq C10 $ \\ 
& $ C10 \sqsubseteq C11 $ \\ 
& $ C0 \sqcap C1 \sqsubseteq C7 $ \\ 
& $ C3 \sqcap C6 \sqsubseteq C8 $ \\ 
& $ C11 \sqcap C12 \sqsubseteq C13 $ \\ 
& $ C1 \sqsubseteq \exists R1.C9 $ \\ 
& $ C3 \sqsubseteq \exists R3.C6 $ \\ 
& $ C9 \sqsubseteq \exists R4.C9 $ \\ 
& $ C9 \sqsubseteq \exists R5.C11 $ \\ 
& $ C9 \sqsubseteq \exists R6.C12 $ \\ 
& $ C9 \sqsubseteq \exists R6.C12 $ \\ 
& $ C10 \sqsubseteq \exists R4.C11 $ \\ 
& $ C10 \sqsubseteq \exists R5.C11 $ \\ 
& $ \exists R2.C3 \sqsubseteq C0 $ \\ 
& $ \exists R3.C7 \sqsubseteq C4 $ \\ 
& $ \exists R4.C10 \sqsubseteq C12 $ \\ 
& $ R4 \sqsubseteq R5 $ \\ 
& $ R2 \sqsubseteq R0 $ \\ 
& $ R4 \circ R6 \sqsubseteq R7 $ \\ 
& $ R3 \circ R0 \sqsubseteq R2 $ \\ 
\hline
\end{tabular}}
\quad \quad
{
\begin{tabular}{|lll|}
\hline
& $ C2 \sqsubseteq C10 $ &\\ 
& $ C2 \sqsubseteq C11 $ &\\ 
& $ C2 \sqsubseteq C12 $ &\\ 
& $ C2 \sqsubseteq C13 $ &\\ 
& $ C4 \sqsubseteq C9 $ &\\ 
& $ C4 \sqsubseteq C10 $ &\\ 
& $ C4 \sqsubseteq C11 $ &\\ 
& $ C4 \sqsubseteq C12 $ &\\ 
& $ C4 \sqsubseteq C13 $ &\\ 
& $ C9 \sqsubseteq C11 $ &\\ 
& $ C9 \sqsubseteq C12 $ &\\ 
& $ C9 \sqsubseteq C13 $ &\\ 
& $ C2 \sqsubseteq \exists R4.C9  $ &\\ 
& $ C2 \sqsubseteq \exists R4.C11  $ &\\ 
& $ C2 \sqsubseteq \exists R5.C9  $ &\\ 
& $ C2 \sqsubseteq \exists R5.C11  $ &\\ 
& $ C2 \sqsubseteq \exists R6.C12  $ &\\ 
& $ C2 \sqsubseteq \exists R7.C12  $ &\\ 
& $ C4 \sqsubseteq \exists R4.C9  $ &\\ 
& $ C4 \sqsubseteq \exists R4.C11  $ &\\ 
& $ C4 \sqsubseteq \exists R5.C9  $ &\\ 
& $ C4 \sqsubseteq \exists R5.C11  $ &\\ 
& $ C4 \sqsubseteq \exists R6.C12  $ &\\ 
& $ C4 \sqsubseteq \exists R7.C12  $ &\\ 
& $ C9 \sqsubseteq \exists R4.C11  $ &\\ 
& $ C9 \sqsubseteq \exists R5.C9  $ &\\ 
& $ C9 \sqsubseteq \exists R7.C12  $ &\\ 
& $  $ &\\ 
& $  $ &\\ 
& $  $ &\\ 
& $  $ &\\ 
& $  $ &\\ 
& $  $ &\\ 
& $  $ &\\ 
& $  $ &\\ 
& $  $ &\\ 
& $  $ &\\ 

\hline
\end{tabular}
}
\caption{\ELp{} Knowledge Graph \& Inference Knowledge Graph}
\label{tab:elp-sample}
\end{table}

\paragraph{RDF Dataset}
For testing the capability of our model in conducting RDF reasoning we are using the same two datasets used in \cite{MakniH19} namely a synthetic
dataset from "Lehigh University Benchmark (LUBM) \cite{DBLP:journals/ws/GuoPH05} and a real-world Scientist dataset from DBpedia. Essentially, the mission for our deep reasoner is to learn the mapping between input RDF
graphs and their inference graphs generated using Apache Jena
API \cite{DBLP:conf/www/CarrollDDRSW04} -a state of the art tool for RDF and OWL reasoning -. 
The first dataset is created on top of LUBM ontology developed for benchmarking Semantic Web knowledge base systems with
respect to use in large OWL applications including deductive reasoning. It conceptualizes 42 classes from the academic domain with 28 properties relating these classes. Using Univ-Bench Artificial Data Generator (UBA) \footnote{http://swat.cse.lehigh.edu/projects/lubm/}, they yielded LUBM1 containing one hundred thousand triples with 17189 subject resources within 15 classes. For each resource $r$ in the set of these subject-resources, a graph
$g$ (graph description of the resource $r$) is created by executing the following SPARQL Query:
\begin{verbatim}
    DESCRIBE <r>
\end{verbatim}

For each knowledge graph g  then they have obtained an inference graph i based on  the LUBM ontology using Apache Jena
API for applying the RDF inference rules covered partially in Table \ref{tab:rdfcomp}.

The second dataset namely "Scientists" is a real-world dataset including $\simeq	 5.5 million$ triples describing 25760 URIs of scientists obtained by applying following SPARQL query against DBpedia \cite{DBLP:conf/semweb/AuerBKLCI07} endpoint:
\begin{verbatim}
prefix dbo: <http://dbpedia.org/ontology/>
select distinct ?scientist
where {
?scientist a dbo:Scientist .
}
\end{verbatim}

The dataset also includes a few other classes related to the Scientist
concept in DBpedia i.e., University and Award related based on set of relationships. For the sake of our evaluations we will conduct our evaluations in each of these classes datasets separately. For a more detailed
description and statistics of these two datasets please see \cite{MakniH19}. 

We split each dataset in a ratio of $80\%-10\%-10\%$ for training, validation, and testing.
\subsection{Training Details}
\label{sec:4-2}
The core of our experiments comprises training a sequence-to-sequence based model trained on large set of knowledge bases and completions pairs $(T,c(T))$. We use two single layer LSTMs of 128 hidden units each: an LSTM encoder for encoding the knowledge graph and and the Pointer LSTM for generating the completion via pointing to the input knowledge base symbols. It has been trained with
stochastic gradient descent, batch size of 100, random uniform weight initialization from -0.08 to 0.08, and L2 gradient clipping of 2.0. The Adam optimizer was used with an initial learning rate of $0.1$. Depending on the maximum knowledge base and completion sizes in our dataset various maximum input sequence lengths and the maximum output lengths have been enforced for each of our experiments.

\subsection{Input Representation Details}
\label{sec:4-3}
\paragraph{Tokenization}For the tokenization of the text we experiment with both Whitespace tokenizers and SubWordText tokenizers \cite{DBLP:conf/acl/SennrichHB16a}. Tokenization of the text is the process of splitting the text into meaningful chunks called tokens. We believe that experimenting with different types of tokenization not only change the accuracy of our results but also gives us better understanding about the nature of reasoning and generalization ability of our network. SubWordText tokeinzer, which works based on variant of byte pair encoding segmentation algorithm \cite{gage1994new}, translates rare words into smaller units than words. As an example, SubWordText tokenizer tokenizes the triples below into \{"http", "www", "department2", "university0", ... \} while the Whitespace tokenizer splits each triple into subject, predicate, and object.   
\begin{verbatim}
<http://www.Department2.University0.edu/GraduateStudent1> 
<http://swat.cse.lehigh.edu/onto/univ-bench.owl#takesCourse>
<http://www.Department2.University0.edu/GraduateCourse0> .
\end{verbatim}
It is worth noting that since the symbols in our synthetic \ELp{} dataset follow the $[A-Z]\backslash d^+$ regular expression both Whitespace and SubWordText tokenizers will lead to the almost same tokens splitting for \ELp{} and hence will not change the results for \ELp{}. 
\paragraph{Normalization}To analyze the deductive reasoning capability of our network as opposed to the inductive reasoning capability usually obtained by learning a good representation of entities during the training or in the pre-training phase here we use the normalization\cite{DBLP:conf/aaaiss/EbrahimiSBXEDKH21,DBLP:conf/iclr/EvansSAKG18}. Unlike inductive reasoning, in the deductive reasoning the names of entities are
insubstantial and should not be leveraged by the reasoner. In \ELp the logical operators ($\sqsubseteq, \exists, \circ, \sqcap, . $) are the only elements of the language in each
knowledge base that have consistent implicit semantics across knowledge bases. In this sense, two entailments
"$A \sqsubseteq C,  \; C \sqsubseteq D \models A \sqsubseteq D$" and  "$P \sqsubseteq Q,  \; Q \sqsubseteq R \models P \sqsubseteq R$" should be treated as equivalent by the ideal reasoner. Similarly for RDF reasoning, 
the actual names (as
strings) of entities from the underlying logic such as variables, constants, functions, and predicates are insubstantial and should not ideally be captured by model. The only elements of the language in each knowledge base that have consistent implicit semantics across the knowledge bases here are the RDF and RDFS controlled vocabulary. Hence, two entailments 	"$(a, \text{rdfs:domain}, x),(y, a, z) \models (y, \text{rdf:type}, x)$" and "$(b, \text{rdfs:domain}, p),(q, b, r) \models (q, \text{rdf:type}, p)$" should be ideally considered as equivalent for logical entailment. Therefore, consistent renaming across a theory should not change the set of entailed formulas (under the same renaming). To encourage models to capture this invariance, we should either provide the term-agnostic input to our model or implement a term-agnostic strategy for the reasoning. To implement the former we use syntactic normalization: a renaming of primitives from the logical symbols to a set of predefined entity
names that are used across different normalized theories.
By randomly assigning the mapping for the renaming, the
network’s learning will be based on the structural information within the theories, and not on the actual names of the
primitives. Note that this normalization not only plays the
role of “forgetting” irrelevant label names, but also makes
it possible to transfer learning from one KB to the other.
Indeed, the network can be trained with many KBs, and
then subsequently tested on completely new ones. To do so, for RDFS reasoning, we normalize all the triples within the knowledge graph by systematically renaming all
URIs which are not in the RDF or RDFS namespaces. Each such URI is mapped to a set of arbitrary strings in a predefined set $A = \{a_1, ..., a_n\}$, where
n is number of entities in our largest KB. Note that URIs in the RDF/RDFS
namespaces are not renamed, as they are important for the
deductive reasoning according to the RDFS model-theoretic
semantics. Consequently, each normalized RDFS KB will
be a collection of facts stored as set of triples $\{(a_i
, a_j , a_k)\}$. Similarly, for the \ELp{} we have generated our syntactic dataset randomly such that only logical operators have consistent semantic meaning across the knowledge graphs. For the latter, later in Section \ref{sec:4-6} we show that - unlike most of the deep learning architectures which mostly rely on learning the symbols' representations - Pointer Network is inherently symbol-invariant and hence we do not need to apply such normalization to the input for Pointer Networks.

\subsection{Baselines}
\label{sec:4-4}
\paragraph{Transformers}
With the recent shift towards using Transformer methods in a variety of tasks and their tremendous success, achieving the state-of-the-art in tasks such as language modeling and machine translation, we believe that it is important to assess their capability in conducting reasoning over the \ELp{} and RDFS deductive reasoning tasks. As such, for our baseline model, we use a standard vanilla encoder-decoder Transformers as proposed in \cite{DBLP:conf/nips/VaswaniSPUJGKP17}. The transformer architecture is merely based on self-attention mechanism and is very parallelizable. It follows with the Encoder-Decoder framework in that, given an input sequence, the network obtain a continuous representation of it based on the context and decode that context-based representation into the output sequence. It replaces the LSTMs with Self-Attention layer and encodes the order using the sinusoidal Positional Encodings. Our network stacks 2 encoder blocks on top of each other where each block consists of 2 sub-layers, a multi-head self-attention mechanism and a position-wise dense feed-forward network. Around each sub-layer a residual connection is employed followed a layer normalization. 
The 2 decoder blocks has the same structure except it contains an additional multi-head attention layer that applied on the output of an encoder block. The multi-head attention that works on output representations masks all subsequent positions and the output embeddings are shifted right by one position so that a prediction for the current step depends only on previously predicted known outputs. 

All embedding layers and all sub-layers in the model produce outputs of size 512. Our Transformer trainer uses the teacher-forcing strategy where the target output gets passed to the next time step regardless of model's prediction at the current time step. The Rectified Linear Unit (ReLU) has been used as our activation function. The batch size has been set to 64 and the multi-head attention consist of 8 heads. The dropout with rate $P_{drop} = 0.1$ has been applied to the output of each sub-layer and also to the embeddings summation and the
positional encodings in both the encoder and decoder stacks.
\paragraph{Graph Words Translation} defines layering RDF graphs for each of the relations in the ontology and encoding them in the form
of 3D adjacency matrices where each layer layout forms a graph word. Each input graph and its entailments are then represented
as sequences of graph words, and RDFS inference can be formulated as sequence-to-sequences problem, solved
using neural machine translation techinques. In an effort to understand the benefits and drawbacks of our method compared to Graph words Translation \cite{MakniH19} -current state-of-the-art method in RDF Reasoning-; here we report our results on the same dataset. Our results show Pointer Networks outperform Graph Words Translation.
\paragraph{Piece-Wise LSTM }
A deductive reasoning involves the learners being given the general rule of entailment in the language, which is then applied to specific knowledge base iteratively. It involves a set of intermediate results added at each step to the original knowledge base until we cannot generate any new statement. The Piece-Wise LSTM and its variants proposed in \cite{DBLP:conf/aaaiss/EberhartEZSH20} strive to emulate this reasoning steps by mapping them to each time step in an LSTM learner. To our knowledge, this is the only work has been done for emulating deductive reasoning for \ELp logic. As such, here we use the same procedure for generating our data and compare our result. Our finding show Pointer Networks outperform Piece-Wise LSTM and its variants by a huge margin.
\paragraph{LSTM Decoder}
As an ablation study, we replace the Pointer LSTM decoder in our encoder-decoder architecture with vanilla LSTM and evaluate its performance. This gives us a clear understanding on the contribution of the Pointer attention in our proposed model.

\subsection{Correctness}
\label{sec:4-5}
In order to reflect how well our Pointer Networks have learned to conduct the reasoning task accurately; here we report the exact matching accuracy for this model and compared that to the above baselines as shown in \ref{tab:correctness-results}. As we can see from the table, our Pointer network model has performed very well (99\% accuracy) in conducting the RDF reasoning outperforming the state-of-the-art results obtained by Graph Words Translation \cite{MakniH19}(98\% accuracy). Later, we show another advantage of Pointer Networks over Graph Words Translation namely its generalization capability to unseen domains. There is also an added benefit that Pointer Networks can be easily applied to any reasoning problems by defining that as "input knowledge base to inferred knowledge base mapping problem". In Graph Words Translation case, however, the network is specifically designed and tailored for RDF reasoning. Compared to the Transformer, our method has shown extraordinarily better accuracy showing clear significant performance gain of Pointer attention over self-attention in conducting the neuro-symbolic reasoning. The very poor performance of our vanilla encoder-decoder LSTM network further corroborate the benefits of using Pointer attentions. Finally, it is worth noting that the random guess is only accurate 2.8e-07\% of the time for RDF reasoning over LUBM dataset demonstrating how difficult this task is.

Similarly, for \ELp{} reasoning task, our proposed model performs extraordinarily well across all the dataset, and achieves much better results achieving 73\% accuracy as opposed to 0.16\% accuracy reported in \cite{DBLP:conf/aaaiss/EberhartEZSH20}. Similarly to our RDF reasoning experiments, here, we found our Pointer attention based method has shown extraordinarily better accuracy compared to the self-attention based Transformers and the vanilla encoder-decoder LSTM network. 

\begin{table}[t]
\centering
\caption{Exact Match Accuracy Results}
\label{tab:correctness-results}
\resizebox{\textwidth}{!}{%
\begin{tabular}{|c|l|l|l||l|l|l|l|}
\hline
\multirow{3}{*}{Logic} & \multirow{3}{*}{KG Size} & \multicolumn{2}{l||}{Pointer Networks}                     & \multicolumn{3}{c|}{Transformer}                                                                                     & \multirow{3}{*}{LSTM} \\ \cline{3-7}
                       &                          & \multirow{2}{*}{SubWordText} & \multirow{2}{*}{Tokenizer} & \multicolumn{1}{c|}{\multirow{2}{*}{Normalized}} & \multicolumn{2}{c|}{Not-Normalized}                               &                       \\ \cline{6-7}
                       &                          &                              &                            & \multicolumn{1}{c|}{}                            & \multicolumn{1}{c|}{SubWordText} & \multicolumn{1}{c|}{Tokenizer} &                       \\ \hline
RDF                    & 3 - 735                  & 87\%                         & \textbf{99\%}                       &                            5\%                      &   25\%                               & 4\%                               & 0.17\%                \\ \hline
\multirow{3}{*}{ER}    & 40                       & 73\%                         & \textbf{73\%}                       & 8\%                                              & 8\%                              & 0.4 \%                         & 0\%                   \\ \cline{2-8} 
                       & 50                       & 68\%                         & \textbf{68\%}                       & 11\%                                             & 11\%                             & 0.3\%                          & 0\%                   \\ \cline{2-8} 
                       & 120                      & 49\%                         & \textbf{49\%}                       &                         15\%                         &                                NA  &                               NA & 0\%                   \\ \hline
\end{tabular}%
}
\end{table}
\subsection{Generalizability: Zero-Shot Reasoning} \label{sec:4-6} 
\begin{table}[t]
\centering
\caption{Exact Match Accuracy Results for Transfer Learning/Representation: SubWordText Tokenization Encoding}
\label{tab:transfer-results-SubWordText}
\begin{tabular}{|l||*{3}{c|}}\hline
\backslashbox{Train}{Test}
&LUBM&Awards&University
\\\hline\hline
LUBM &*&75\%&78\%\\\hline
Awards &79\%&*&77\%\\\hline
University &81\%&82\%&*\\\hline
\end{tabular}
\end{table}

\begin{table}[ht]
\centering
\caption{Exact Match Accuracy Results for Transfer Learning/ Representation: Whitespace Tokenization Encoding}
\label{tab:transfer-results-Whitespace}
\begin{tabular}{|l||*{3}{c|}}\hline
\backslashbox{Train}{Test}
&LUBM&Awards&University
\\\hline\hline
LUBM &*&61\%&47\%\\\hline
Awards &96\%&*&84\%\\\hline
University &82\%&88\%&*\\\hline
\end{tabular}
\end{table}

Using the pointer networks for simply copying from the input knowledge base to the completed one might seem simple. Despite their simple nature, the generalizability that they can provide is far more intricate and in our interest. Indeed, the main goal of this paper is to demonstrate the general symbol/naming-invariant reasoning learning capability of Pointer Networks when they encounter with the knowledge graph of the new domain/vocabulary in the testing phase. This ensures the model has gained the deep understanding of the logical semantics and reasoning as opposed to merely working based on the representation learning and induction. As such, we measured exact matching accuracy of the results yielded by Pointer Networks when trained on one domain and tested on another without fine-tuning i.e., zero-shot . The results for its transfer capability for RDFS reasoning is shown in Tables \ref{tab:transfer-results-SubWordText} and \ref{tab:transfer-results-Whitespace}, while for the \ELp{} the results in Table \ref{tab:correctness-results} already shown this capability. Based on the table, Pointer Network gives surprisingly good and consistent empirical results when it comes to transfer learning. Indeed, Pointer Networks have several desirable inherent characteristics leading into this transfer learning behaviour. They are capable of dealing with 
dynamic vocabulary length as opposed to fixed vocabulary output, dealing with 
rare or out-of-vocabulary word, and
heavy-tailed vocabulary distribution. 

Additionally, to further understand the nature of how Transformers learn to reason?, we have applied normalization and various tokenization on our RDF dataset and examined the change in the accuracy. Not surprisingly, unlike Pointer Networks, Transformers are very sensitive to the changes of tokenization and the normalization. This is mainly because Transformers heavily rely on the subsymbolic representation
of entities and relations learned by the network. Indeed the power of Transformers mainly comes from their self/intra-attention module primarily used to learn the representation of the tokens in the input based on their relations with other tokens. This explains why our baseline Transformer model tends to obtain its highest accuracy when trained on not-normalized SubWordText encoded RDF knowledge base. This way, the network can learn much better representation for the symbols in the knowledge base which leads to better reasoning accuracy. Unsurprisingly, the normalization decreases the accuracy of the Transformer showing poor symbol-invariant reasoning capability, as indicated in the \textit{Normalized} column in Table \ref{tab:correctness-results}.

\section{Conclusion \& Future Works}
\label{sec:5}
We have shown that a deep learning architecture based on pointer networks is capable of learning how to perform deductive reason over RDFS and \ELp\ KBs with high accuracy. We designed a novel way of conducting neuro-symbolic through pointing to the input elements. More importantly we showed that the proposed approach is generalizable across new domain and vocabulary demonstrating symbol-invariant zero-shot reasoning capability. We plan to properly investigate scalability of our approach and to adapt it to other, more complex logics. We furthermore intend to investigate the added values which should arise out of adding subsymbolic deductive reasoning components to more traditional deep learning scenarios, in particular in the areas of knowledge graph inference and natural-language-based commonsense reasoning.


%

\bibliographystyle{unsrtnat}
\bibliography{references}  






\end{document}